**Improving VAE based molecular representations for compound property prediction**

Ani Tevosyan[a], , Lusine Khondkaryan[b], Hrant Khachatrian[a,] , Gohar Tadevosyan[b], Lilit Apresyan[b], Nelly Babayan[b], Helga Stopper[c], Zaven Navoyan[d]*

[a] YerevaNN, Charents str. 20, 0025, Yerevan, Armenia

[b] Group of Cell Technologies, Institute of Molecular Biology, National Academy of Sciences of RA, Hasratyan str. 7, 0014, Yerevan, Armenia

[c] Department of Toxicology, Institute of Pharmacology and Toxicology, University of Würzburg, Versbacher str. 9,97078, Würzburg, Germany

[d] Toxometris.ai, Armenia, Hasratyan str. 7, 0014, Yerevan, Armenia,

[*] **Corresponding author**: Zaven Navoyan, Toxometris.ai, Armenia, Hasratyan str. 7, 0014, Yerevan, Armenia

**E-mail address**: znavoyan@gmail.com

## Abstract

Collecting labeled data for many important tasks in chemoinformatics is time consuming and requires expensive experiments. In recent years, machine learning has been used to learn rich representations of molecules using large scale unlabeled molecular datasets and transfer the knowledge to solve the more challenging tasks with limited datasets. Variational autoencoders are one of the tools that have been proposed to perform the transfer for both chemical property prediction and molecular generation tasks. In this work we propose a simple method to improve chemical property prediction performance of machine learning models by incorporating additional information on correlated molecular descriptors in the representations learned by variational autoencoders. We verify the method on three property prediction tasks. We explore the impact of the number of incorporated descriptors, correlation between the descriptors and the target properties, sizes of the datasets etc. Finally, we show the relation between the performance of property prediction models and the distance between property prediction dataset and the larger unlabeled dataset in the representation space.

## Introduction

One of the challenging tasks in chemoinformatics and molecular modeling is the prediction of physicochemical properties and/or biological activities of chemical compounds from their molecular/chemical features. Quantitative structure-activity/property relationship (QSAR/QSPR) modeling based on machine learning (ML) methods, though, achieved a certain amount of success [1,2,3], but still suffer from a number of issues. Prediction efficacy and performance of ML methods is often limited by the availability of labeled datasets. The size of the training data is acknowledged as a key property for reliable and accurate predictions [4,5]. Another obstacle of classical machine learning is reliance on similarity of chemical/biological space occupied by training and test datasets [6,7]. Difference in training and test data distribution may lead to faulty extrapolation of the model and even degradation [8]. This scenario implies collection of a new training dataset, which in practice is hard to achieve due to expensive and time-consuming experimental procedures. One of the proposed solutions is called transfer learning, which consists of model pre-training on a large scale dataset with cheap or free labels and a further fine-tuning on a smaller target dataset [9,10]. Ideally, this pre-training causes the model to develop general-purpose abilities and knowledge that can then be transferred to downstream tasks. Recent developments in transfer learning techniques have shown competitive results in QSAR/QSPR modeling [11,12,13]. Additionally, the performance of an ML algorithm depends to a large extent on the way input data is represented. Consequently, various descriptors and fingerprints engineered by experts have been widely employed for properties predictions [14,15].

Achievements in deep neural networks put forward data-driven representation learning based on molecular graphs [16,17,18] or sequence-based strings, like SMILES (Simplified Molecular Input Line Entry Specification) [19,20]. In recent years, unsupervised latent representation of molecules extracted from variational autoencoders (VAEs) with encoder-decoder architecture for QSAR/QSPR tasks has gained significant attention. One of the earliest works in this area is chemical VAE (CVAE) proposed by Gómez-Bombarelli *et al.* [21], which converts a SMILES sequence to and from a fixed sized continuous vectors. Though the main purpose of the developed version of VAE was de *novo* molecule generation, the authors also utilized latent representations for property predictions. Joint training of VAE with a property predictor results in a latent space organized in a way that the molecules are ordered according to their property values. Since then, various adaptations and modifications of VAEs have been proposed, aiming to improve latent space organization [22,23,24,25,26]. Despite huge efforts, the performance of VAEs is still far from perfect. Recently, latent representations extracted from VAEs pre-trained on large datasets have been exploited as a basis for transfer learning [27,28]. It was shown that although this approach resulted in reasonable performance, simpler ML models like gradient boosting or graph convolutional networks without transfer learning still outperform it.

In the present work, we aim to test the hypothesis that jointly training a VAE with an additional predictor of descriptors specifically correlated with the property of interest may impact the latent space quality and, consequently, improve the prediction accuracy of the downstream task. For this, we adopted the following approach (Fig.1): first we identified a subset of descriptors mostly correlated with a particular target in downstream tasks, then we pre-trained a VAE along with a predictor of the chosen descriptors on the data-rich ZINC dataset. Afterwards, extracted embeddings were utilized in building models for the target regression and classification tasks.

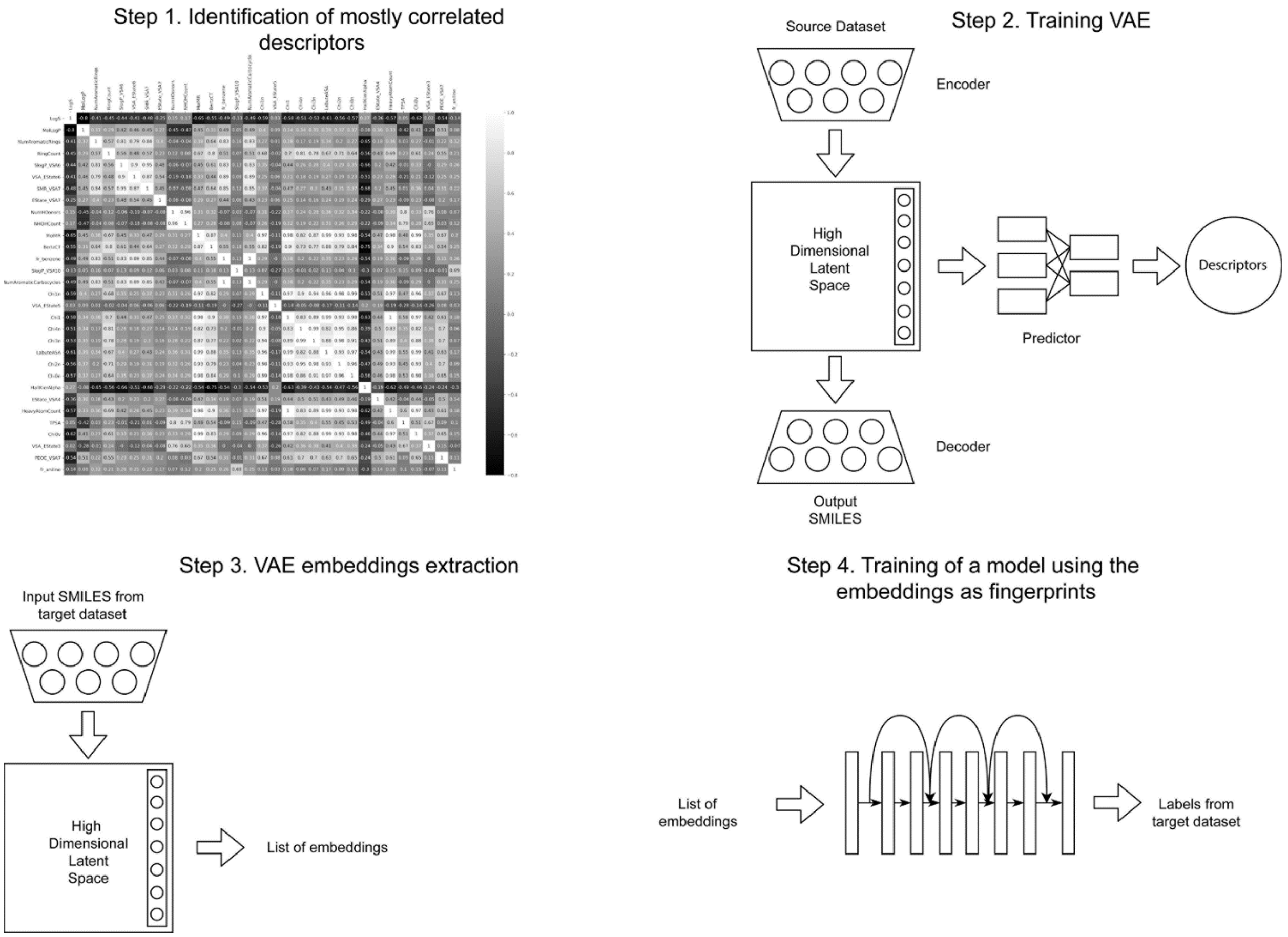

**Fig. 1** The training workflow

## Methods

### *Datasets and preprocessing*

Throughout the paper, the term “source dataset” refers to a large unlabeled dataset of small molecules which can be utilized in learning molecular representations. While the source dataset is unlabeled, we can still compute values of many descriptors for all the molecules using chemical software libraries. The “target dataset” refers to a small-scale labeled dataset for a particular downstream task which will be used in QSAR/QSPR modeling.

Following previous studies [21,22], we have used a ZINC-250K as a source dataset. It consists of 250000 drug-like molecules downloaded from the ZINC15 database [29].

The proposed approach was tested in two regression and one classification QSAR/QSPR problems. Table 1 shows the sizes of relevant target datasets. Aqueous solubility (logS) data were adopted from Cui et al. [30], which was collected from ChemIDplus database and literature search. The experimental lipophilicity (logD) data were retrieved from ChEMBL (version 29). The data was curated so that only results for octanol/aqueous buffer determined by conventional shake-flask method at pH 7.4 were kept. Blood-brain barrier penetration (logBB) data was obtained and integrated from various studies [31,32,33,34,35,36,37]. For the initial data processing, the duplicates from all datasets were removed by InChiKeys comparisons, using RDKit [38]. Compounds with equivocal values or contradicting data were discarded. Additionally, all salts and mixtures were excluded. Furthermore, since most of the collected logBB data were divided into two classes, we applied the same threshold used in the sourced publications to classify compounds into BBB+ and BBB- classes based on the criterion logBB≥-1 or logBB<-1, respectively. We note that logBB prediction is significantly harder than the other two tasks as it involves complex biological processes. The datasets are accessible from our github repository (https://github.com/znavoyan/vae-embeddings).

**Table 1** Description of QSAR/QSPR datasets

| Dataset | Acronym | Task | Final dataset size |
|---|---|---|---|
| Aqueous solubility | logS | regression | 6668 |
| Molecular lipophilicity | logD | regression | 6217 |
| Blood-brain barrier penetration | logBB | classification | 2906 |

*Descriptors selection*

For each of the target datasets, we have computed molecular descriptors for all compounds via RDkit package [38]. The computed data was inspected for outliers and missing values, which were dropped off from the dataset. A variance threshold of 0.5 was used to remove all descriptors with low variance. Afterwards, the optimal subset of descriptors for each of the tasks was determined by the filter-based linear correlation method [39]. Note that for the logBB dataset, the correlation was computed only for the subset of compounds for which we had numeric values. The intercorrelated descriptors with the Pearson correlation coefficient r>0.9 were eliminated by excluding one of the descriptors (correlation heatmaps are presented in Additional file 1: Fig. S1). Finally, three descriptors mostly relevant with a target task were selected (Table 2).

**Table 2** Descriptors used to train VAEs for different downstream problems

| Dataset | Selected descriptors | Pearson correlation coefficient (r) |
|---|---|---|
| Aqueous solubility (logS) | MolLogP | -0.8 |
| | PEOE_VSA6 | -0.61 |
| | MolWt | -0.59 |
| Molecular lipophilicity (logD) | MolLogP | 0.43 |
| | NumAromaticRings | 0.29 |
| | RingCount | 0.26 |
| | TPSA | -0.57 |
| Blood-brain barrier penetration (logBB) | NumHAcceptors | -0.43 |
| | NumHeteroatoms | -0.43 |

*Variational autoencoders*

Variational autoencoders (VAEs) are generative models designed to model an unknown data distribution using a finite sample from the distribution. A regular VAE consists of an encoder which maps every input sample to a normal distribution over the continuous latent space, and a decoder which maps every point of the latent space to a point in the input space [40]. Encoder and decoder are defined as neural networks and are jointly trained to optimize a lower bound on the likelihood of the input samples. In practice it is implemented by minimization of a loss function with two terms: a reconstruction term, which forces the decoder to correctly reconstruct the input given its latent representation, and a regularization term which is the Kullback-Leibler (KL) divergence between the conditional distributions given by the encoder and some predefined prior distribution.

Chemical VAE (CVAE) adapts this architecture for small molecules [21]. The inputs to the VAEs are small molecules represented by a SMILES string. CVAE also adds an additional predictor module on top of the latent representation which can be used to predict a set of properties for the input molecules. The predictor is implemented as a multi-layer perceptron (MLP) with 3 hidden layers.

Penalized VAE (PVAE) applies weights on the output alphabet of the decoder depending on the prevalence of characters in SMILES strings [26]. It helps to significantly improve the reconstruction accuracy of the model. CVAE uses a convolutional network for the encoder and Gated recurrent units (GRU) [41] for the decoder. PVAE uses GRU networks for both, and a single linear layer for the predictor. We use both CVAE and PVAE in our experiments. The latent space is 196-dimensional, hence each molecule will be represented by 196-dimensional vectors in the downstream tasks. The hyperparameters of VAEs used in this work are presented in Additional file 1: Table S1.

*QSAR models*

Three ML models were applied on latent representations: Multilayer Perceptron (MLP), Linear Regression (LR) and deep neural network based on 1D ResNet (residual network) architecture. The first two are conventional algorithms in ML, while the usage of 1D ResNet in prediction of compound's physico-chemical properties have been proposed by Cui et al. [30] and showed competitive results in aqueous solubility prediction. Since the source code for 1D ResNet was not available, we have re-implemented the architecture by following the instructions in the paper. The models were optimized with respect to their hyperparameters (Additional file 1: Table S2). The training was done using the 10-fold cross validation method with the scikit-learn library for Python [42].

*Evaluation metrics and distance measurements*

The predictive performance of the models was evaluated by the coefficient of determination ($R^2$) and root mean squared error (RMSE) for regression tasks, while accuracy and F1(harmonic mean of the precision and recall) was used for classification task.

Additionally we were interested in the performance of the models on parts of the target dataset with respect to their distance from the source dataset. Formally, we clustered the target dataset using K-Means algorithm, and calculated the distance between the cluster and the source dataset.

There are many ways to compute the distance between two sets of molecules. Recently K. Preuer et al. proposed Fréchet ChemNet Distance [43]. It assumes that the distributions of the molecules are Gaussian in the latent space, estimates the means and covariance matrices of the Gaussians, and calculates the Fréchet distance between them. We believe this assumption is unrealistic and suggest measuring the distance using the tools provided by VAEs. The encoder of a VAE maps every input molecule to a Gaussian distribution in the latent space parametrized by the encoder's outputs. VAEs minimize the expected distance between these Gaussians to the prior distribution measured by KL divergence. We suggest using the same expected KL divergence to measure the distance between target dataset and the prior, and compare it to the distance between source dataset and the prior. Note that unlike Fréchet ChemNet Distance, this definition of the distance between distributions in the latent space is applicable only to VAE-based models.

## Results and discussion

### *The influence of the correlated descriptors*

In order to investigate the influence of VAE pre-training with relevant descriptors on the quality of latent space, a set of experiments were carried out. First, both types of VAEs were pre-trained with the most correlated task-specific descriptor on the ZINC-250K source dataset. As a baseline, we used VAEs pre-trained without the predictor modules. The extracted embeddings were utilized in the three target tasks.

Jointly pre-training with the most relevant property improved downstream prediction performance across both types of VAEs, three QSAR models and three downstream tasks, except for PVAE on logBB prediction task (Table 3 and 4). This trend was even more evident in the case of the CVAE.

The pattern was observed even for a relatively weak correlation seen between logD and MolLogP (r=-0.43).

The results also indicate that PVAE outperformed the CVAE in all three tasks. Among the studied QSAR models, the best performance was achieved with 1D ResNet, followed by MLP. In logBB prediction tasks, linear regression and MLP gave similar results.

**Table 3** Performance of various models for predicting logS and logD tasks

| Task | VAE | Descriptor | Pearson correlation coefficient (r) | Classifier | | |
|---|---|---|---|---|---|---|
| | | | | 1D ResNet | MLP | LR |
| | | | | $R^2$/RMSE | $R^2$/RMSE | $R^2$/RMSE |
| logS | CVAE | - | - | 0.568/1.334 | 0.557/1.349 | 0.452/1.502 |
| | CVAE | MolLogP | -0.8 | 0.772/0.966 | 0.764/0.984 | **0.760/0.994** |
| | PVAE | - | - | 0.745/1.022 | 0.668/1.165 | 0.641/1.212 |
| | PVAE | MolLogP | -0.8 | **0.796/0.913** | **0.770/0.971** | 0.753/1.005 |
| logD | CVAE | - | | 0.313/ 1.007 | 0.185/ 1.097 | 0.159/ 1.115 |
| | CVAE | MolLogP | -0.43 | 0.397/ 0.943 | 0.292/ 1.023 | 0.276/ 1.034 |
| | PVAE | - | | 0.434/ 0.913 | 0.242/ 1.057 | 0.207/ 1.081 |
| | PVAE | MolLogP | -0.43 | **0.520/ 0.840** | **0.319 / 1.001** | **0.296 / 1.018** |

Bold values indicate the best performance over all models

**Table 4** Accuracy and F1 score of various models for predicting logBB task

| Task | VAE | Descriptor | Pearson correlation coefficient (r) | Classifier | | |
|---|---|---|---|---|---|---|
| | | | | 1D ResNet | MLP | LR |
| | | | | Accuracy/F1 | Accuracy/F1 | Accuracy/F1 |
| logBB | CVAE | - | - | 0.842/0.906 | 0.843/0.907 | 0.838/0.906 |
| | CVAE | TPSA | -0.57 | 0.865/0.921 | 0.862/0.917 | 0.867/0.922 |
| | PVAE | - | - | 0.887/0.932 | 0.874/0.927 | 0.877/0.929 |

| | | | | | |
|---|---|---|---|---|---|
| PVAE | TPSA | -0.57 | **0.888/0.934** | **0.876/0.928** | **0.880/0.931** |

Bold values indicate the best performance over all models

While the logBB prediction task is the hardest task we examined, we attempt to analyse what factors could explain the poor influence of the joint pre-training with the correlated descriptor in case of PVAE. One possible explanation is that the molecule's structural features and the chemical property, i.e. topological surface area (TPSA), are already embedded in the PVAE space even without explicitly training it with the TPSA predictor. To verify this hypothesis, we measured how much information about the TPSA descriptor is contained in the embeddings by training a linear model on top of the embeddings to predict TPSA. Table 5 shows that although adding a TPSA predictor to PVAE improves TPSA predicting performance of PVAE embeddings, the difference is much smaller than in CVAE. Indeed, the raw PVAE has more information about TPSA than the raw CVAE.

**Table 5** Topological surface area (TPSA) prediction performance measured by root mean square error (RMSE) of four VAE models

| VAE | Descriptor | TPSA prediction RMSE |
|---|---|---|
| CVAE | - | 0.643 ± 0.048 |
| CVAE | TPSA | 0.066 ± 0.011 |
| PVAE | - | 0.390 ± 0.023 |
| PVAE | TPSA | 0.136 ± 0.011 |

Another difference between logS and logBB prediction tasks is the size of the target datasets. Table 6 shows that if we train a ResNet model to predict logS on a dataset with the same size as the logBB dataset, still the version of PVAE with a descriptor predictor performs better. So while the size of the target dataset is important, it does not explain the poor influence of the TPSA predictor.

**Table 6** Prediction performance of logS and logBB tasks when the training data has the same number of samples

| | Dataset size | No descriptors | With the top 1 descriptor |
|---|---|---|---|
| logS ($R^2$) | 6668 | 0.745 | 0.796 |
| logS ($R^2$) | 2906 | 0.676 | 0.763 |
| logBB (Accuracy/F1) | 2906 | 0.887/0.932 | 0.888/0.934 |

Finally, we note that our logBB dataset contains compounds that penetrate blood-brain barrier via both passive diffusion and active transport (i.e. through "structure-specific" transporters) mechanisms, which makes structure based information even more relevant than property-based one [44].

*Variance of the models*

In our experiments we have noticed a high variance in the performance of VAE-based QSAR models. In this section we discuss the sources of the observed variance in detail. Our experimental setup induces four such sources:

1. Pre-training of the VAE, which comes from the initialization of its weights and the stochastisticity of the training batches,
2. The noise induced by VAE's encoder when obtaining the embeddings of molecules,
3. Training / test split of the target dataset,
4. Training of the downstream QSAR/QSPR model.

To save time and compute, we focus on linear regression models in this section. They are trained in a deterministic fashion and hence eliminate the fourth source of the variance.

We obtain embeddings of the molecules by passing the entire target dataset through the encoder of a pre-trained PVAE (with MolLogP) five times, split the dataset into 10 folds and train linear regression models to predict the target property. The results are displayed on Fig. 2. While the standard deviation of the $R^2$ of the models across the folds is 0.023 and 0.034 for logS and logD, respectively (on average), we report the mean $R^2$ of the 10 folds, and the standard deviation of these means across different embeddings is only 0.001 and 0.003 for logS and logD, respectively.

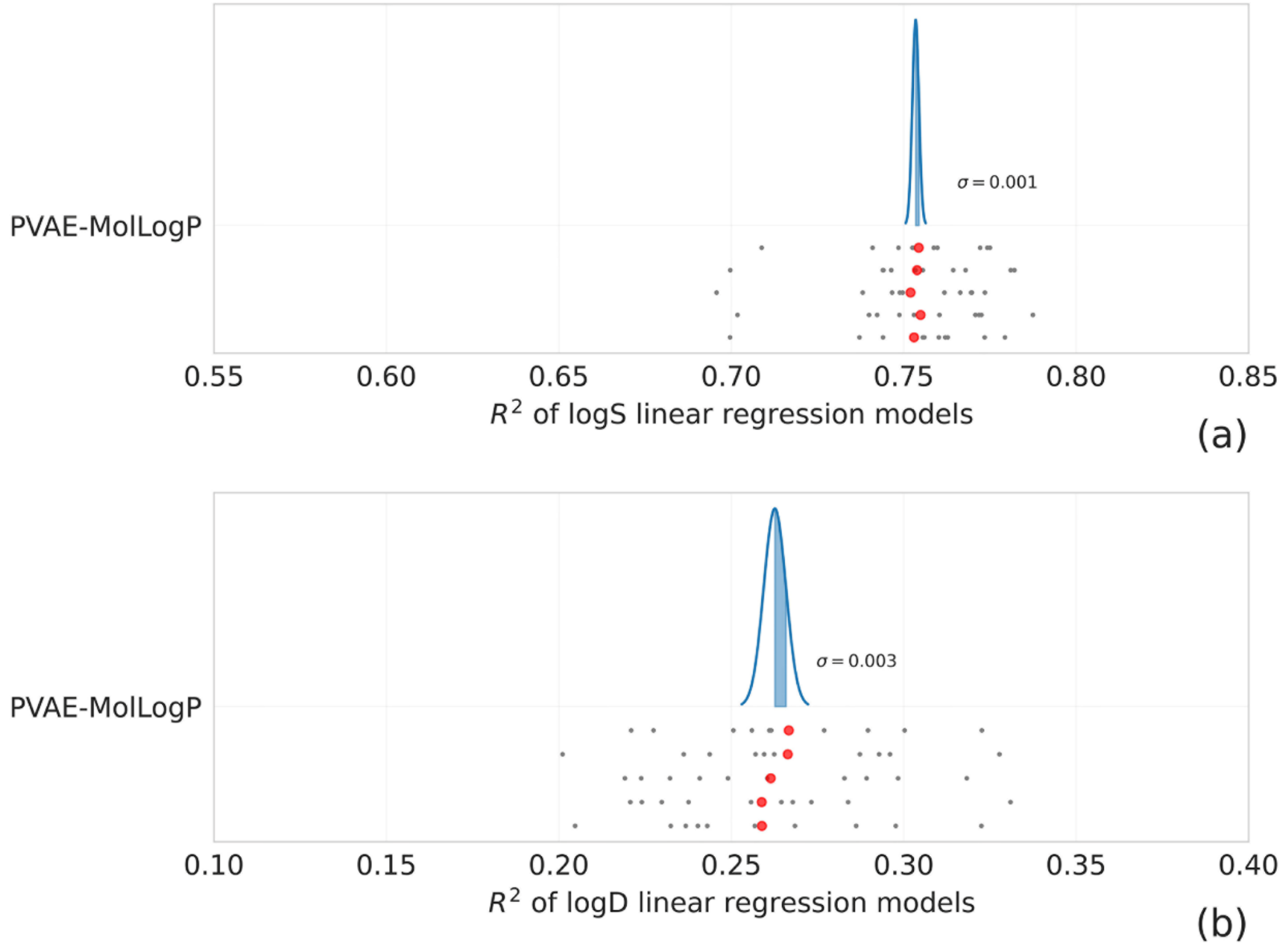

**Fig 2** Means and variances of $R^2$ metrics of linear models over 5 different embeddings obtained from a PVAE trained with a MolLogP predictor. The gray points represent the $R^2$ value of a fold, while the red points show the mean of 10 folds. The mean and variance of the Bell curve is calculated based on the values of red points. Curves on (a) and (b) graphs represent logS and logD downstream tasks correspondingly

To measure the variance coming from the first source, we train five PVAEs of two types: without a predictor, and with a MolLogP-predictor. Then we train linear regression models on the representations taken from each of these VAEs (only one embedding per VAE) and visualize the evaluation metrics. As always, we perform 10-fold cross-validation, and average the scores across the folds. The Bell curves in Fig. 3 represent normal distributions computed by the mean and the variance of five scores, one per each PVAE. The variance coming from the pre-training of VAEs is much larger than the variance coming from the encoder's outputs.

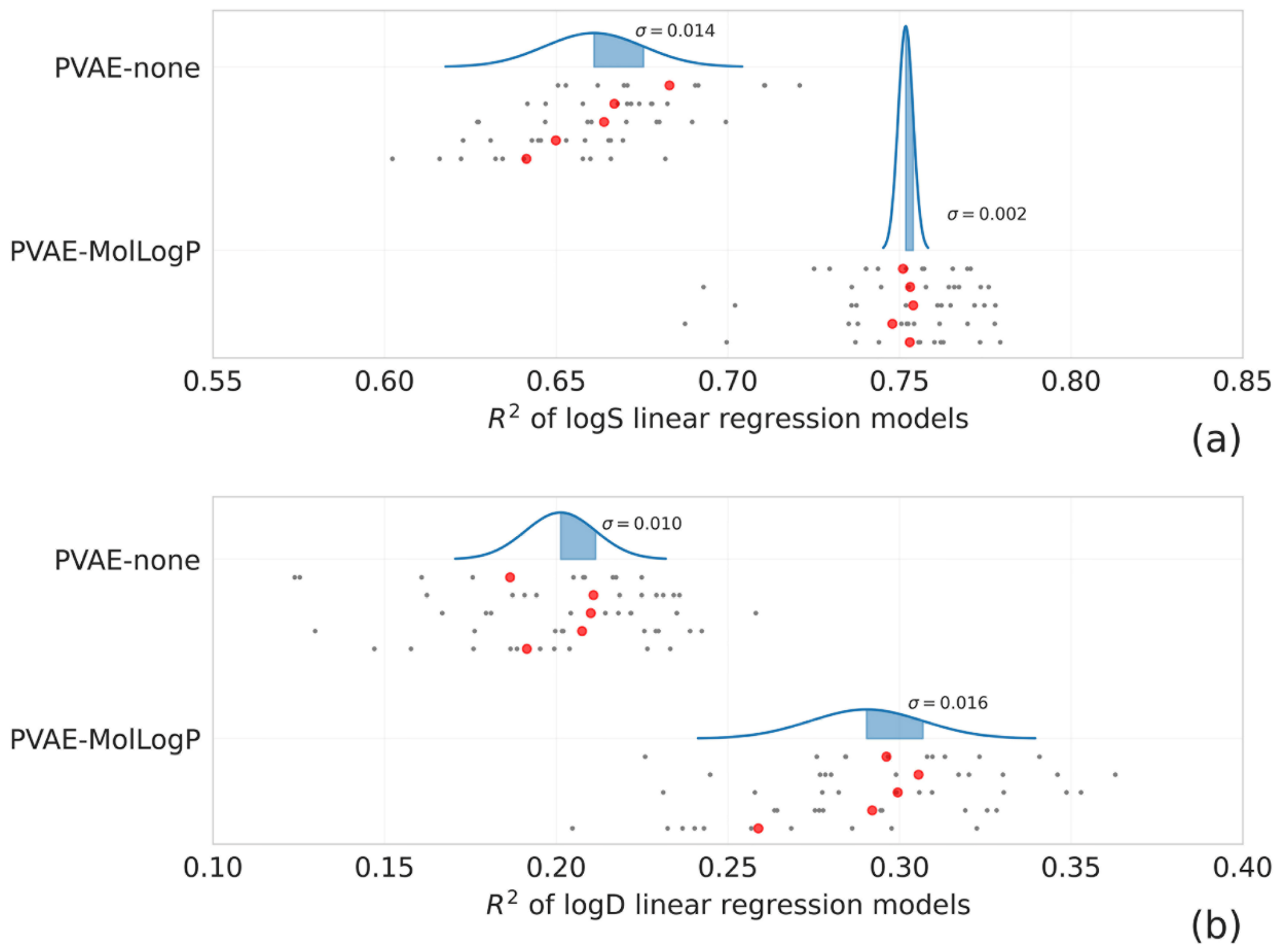

**Fig. 3** Means and variances over 5 VAEs trained with and without predictor are presented in the form of Bell curve. The gray points represent the R2 value of a single fold, while the red points show the mean of the scores of 10 folds. The mean and variance of the Bell curve is calculated based on the values of red points. Curves on (a) and (b) graphs represent logS and logD prediction tasks, respectively

*Leveraging multiple correlated descriptors*

Since better results were achieved with the PVAE, it was used in all further experiments. To push the strategy of incorporating descriptor information into VAEs to its limit, we tried to arrange the chemical space by a subset of descriptors. For this, the PVAE was pre-trained with the top two and top three descriptors correlated with the target variable. The behavior varied a lot across the setups. As evident from the results (Table 7), adding the second most correlated descriptor improved logS prediction performance for the ResNet model, while for linear regression, adding the third one was necessary to see a gain. For MLP, further addition of the descriptors did not lead to significant changes. In case of logD prediction, addition of the second descriptor helped only in MLP model, while two other models got worse with the additional descriptors. In logBB prediction adding new descriptors does not significantly affect the performance (Table 8).

**Table 7** Prediction performance of logS and logD tasks based on embedding extracted from VAEs jointly trained with several descriptors

| **Task** | **Descriptors** | **Classifier** | | |
|---|---|---|---|---|
| | | **1D ResNet** | **MLP** | **LR** |
| | | **$R^2$/RMSE** | **$R^2$/RMSE** | **$R^2$/RMSE** |
| logS | MolLogP | 0.790/0.926 | 0.770/0.971 | 0.751/1.010 |
| | MolLogP, PEOE_VSA6 | **0.809/0.88** | 0.769/0.970 | 0.752/1.006 |
| | logP, PEOE_VSA6, MolWt | 0.804/0.896 | 0.771/0.967 | **0.762/0.986** |
| logD | MolLogP | 0.520 / 0.840 | 0.319 / 1.001 | **0.296 / 1.018** |
| | MolLogP, NumAromaticRings | **0.522 / 0.839** | **0.335/0.989** | 0.295/1.019 |
| | MolLogP, NumAromaticRings, | 0.510/ 0.851 | 0.303/1.013 | 0.277/1.032 |

| | RingCount | | | |
|---|---|---|---|---|

Bold values indicate the best performance over all models

**Table 8** Prediction performance of logBB task based on embedding extracted from VAEs jointly trained with several descriptors

| Task | Descriptors | Classifier<br>1D ResNet<br>Accuracy/F1 | <br>MLP<br>Accuracy/F1 | <br>LR<br>Accuracy/F1 |
|---|---|---|---|---|
| logBB | TPSA | 0.888/0.934 | 0.876/0.928 | 0.880/0.931 |
| | TPSA, NumHAcceptors | 0.886/0.933 | **0.878/0.929** | 0.880/0.930 |
| | TPSA, NumHAcceptors, NumHeteroatoms | **0.891/0.935** | 0.875/0.928 | **0.882/0.931** |

Bold values indicate the best performance over all models

To explain why the pre-training with the top three correlated descriptors gives opposite effects for the linear regression model of logS and logD prediction tasks, we hypothesize that PVAE struggles to smoothly organize the latent space with multiple descriptors in case of the logD-correlated descriptors, but not in case of the logS-correlated descriptors. One way to measure it is to see whether the representations retain information about MolLogP when more descriptors are used in the pre-training phase. We trained linear regression models to predict MolLogP from various representations and calculated their RMSE errors. Table 9 shows that logD-correlated descriptors do not reduce the quality of representations in terms of the information kept about MolLogP, so this experiment did not support our hypothesis.

**Table 9** The MolLogP prediction power

| Descriptors | MolLogP RMSE (logS-correlated descriptors) | MolLogP RMSE (logD-correlated descriptors) |
|---|---|---|
| MolLogP | 0.287 ± 0.0009 | 0.287 ± 0.0009 |
| MolLogP + 1 more | 0.290 ± 0.0010 | 0.285 ± 0.0008 |
| MolLogP + 2 more | 0.290 ± 0.0006 | 0.285 ± 0.0008 |

Previously, it has been suggested that multi task-learning neural networks can benefit only from training with correlated features, otherwise causing model degradation [45,46]. Our results indicate that this is not always the case. The embeddings extracted from the PVAE jointly pre-trained with multiple descriptors with low intercorrelation were able to achieve predictive power comparable to single-task pre-training in many cases, while in other cases they achieved a noticeable stronger performance.

*The impact of the correlation coefficient*

To evaluate the association between predictive performance and the correlation between the target variable and the descriptor, we pre-trained PVAE with several descriptors with descending correlation coefficients. The predictive performance of the extracted embedding was evaluated for logS and logD datasets using ResNet models. We expected the performance to decrease when pre-training with less correlated descriptors. Additionally, we tried to verify whether the correlation was the only factor affecting the downstream performance. To do that we have created “synthetic” descriptors with lower correlation to the target variable by adding random noise to the MolLogP descriptor. The noise reduced the descriptor’s correlation to match the corresponding values of the selected real descriptors.

As expected, predictive performance of the models generally worsened as the correlation coefficient decreased, both in case of real descriptors and synthetic descriptors with noisy MolLogP (Fig. 4). We have identified a few exceptions. In particular, logS prediction performance (Fig. 4a) was better when the PVAE was pre-trained with MolWt (r=0.59) compared to the PVAE pre-trained with PEOE_VSA6 (r=0.61). This is the first evidence that while the correlation coefficient is important, it is not the sole factor. The other factors might be connected to the underlying physics of the particular descriptors. The other, stronger evidence comes from the experiments with the synthetic descriptors. In 7 out of 8 experiments, the PVAEs pre-trained with noisy MolLogP predictors performed better than the PVAEs pre-trained with descriptors having the same correlation coefficient. In general, the mentioned patterns are less expressed on logD prediction tasks (Fig. 4b) possibly due to the higher variance of the performance of VAE-based models (Section 3.2).

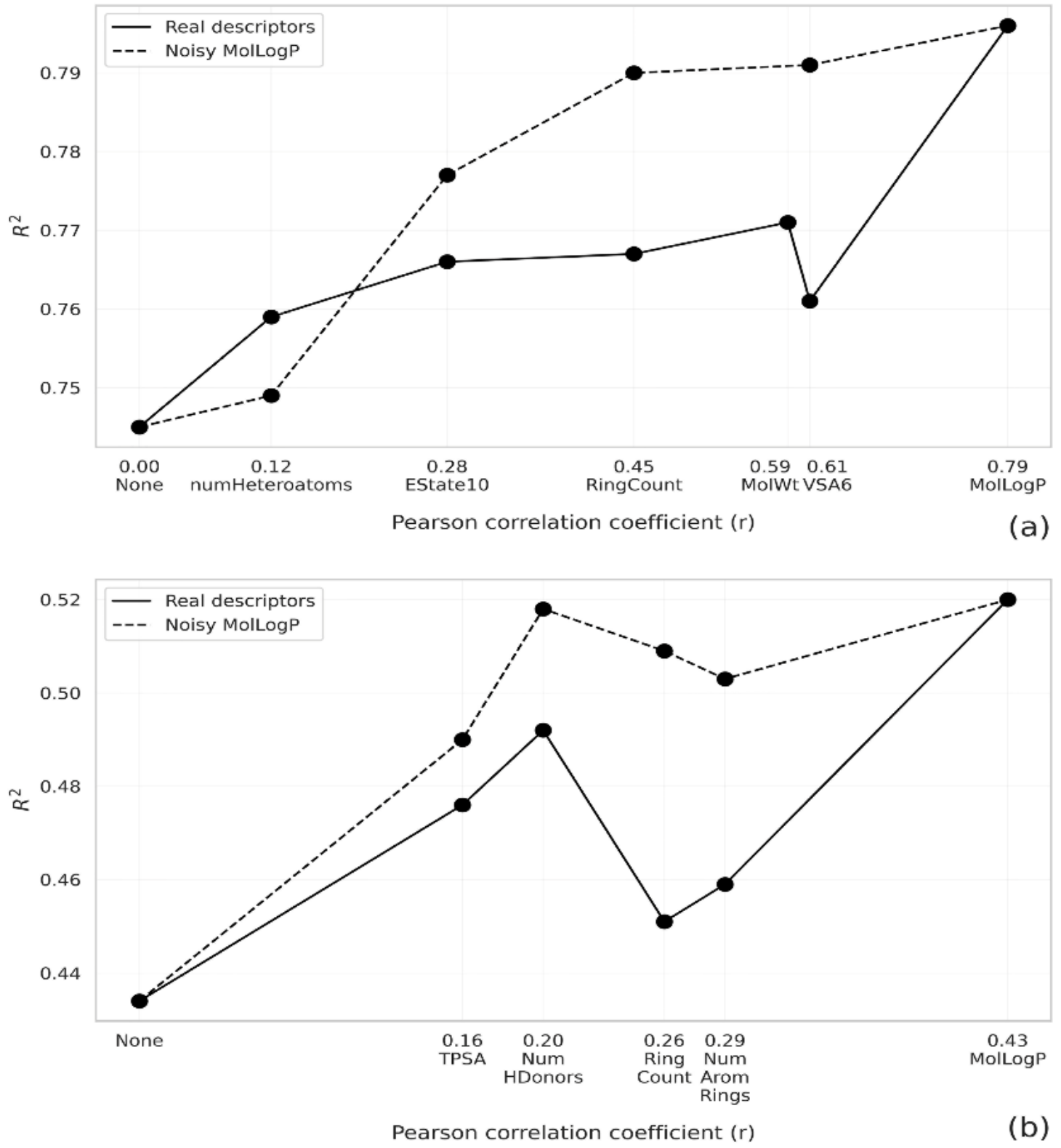

**Fig. 4** Predictive performance of 1D ResNet models as a function of the correlation coefficients between descriptors and the target variables. (a) logS prediction task, (b) logD prediction task

*The impact of the source dataset's size*

Furthermore, we aimed to assess how the size of the source dataset affects the reconstruction accuracy and prediction quality in a downstream task. Here, two versions of PVAE, with and without joint training with a descriptor, were pre-trained on the random subsets of ZINC-250K dataset, with the sizes ranging between 2% and 100% of the original size. Molecular embedding extracted from the pre-trained PVAEs were utilized in the logS prediction task. As can be seen from Fig. 5a, in both setups of PVAE pre-training, the reconstruction accuracy gradually increased from 2.5% up to 76.8% when the size of the source dataset was increased from 5K to 250K. In contrast, predictive performance reached the plateau at the 25K and further increase of the source training set led to only slight perturbation in models predictive power (Fig. 5b). The similar pattern was preserved in case of joint pre-training.

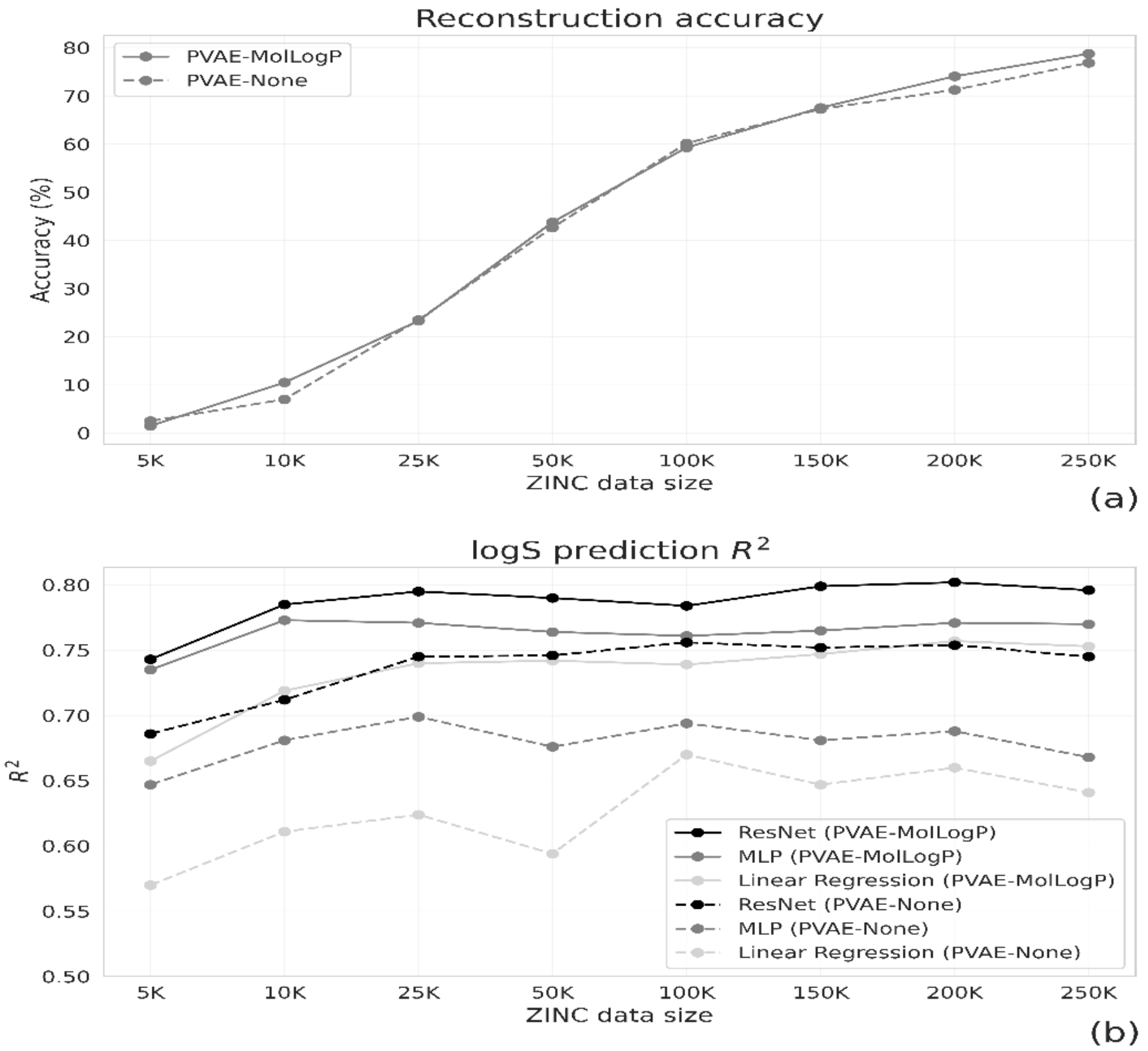

**Fig. 5** Reconstruction accuracy (a) and logS prediction quality (b) of PVAE models pre-trained on subsets of ZINC-250K

*Latent space structure*

First we visualize how the added property affects the order of molecules in the latent space. For that, we forward pass the test set of the ZINC-250K source dataset through the encoders of pre-trained VAEs, perform a 2-dimensional PCA on the 196-dimensional vectors and color them by the value of the respective descriptor. The results are shown on Fig. 6. Adding the descriptor always makes similarly colored points to get closer, similar to what was observed by Gómez-Bombarelli et al. [21] On the other hand, for the VAEs trained with and without TPSA we see a large difference between CVAE models and a smaller difference between PVAE models, which is consistent with the results seen in Table 5.

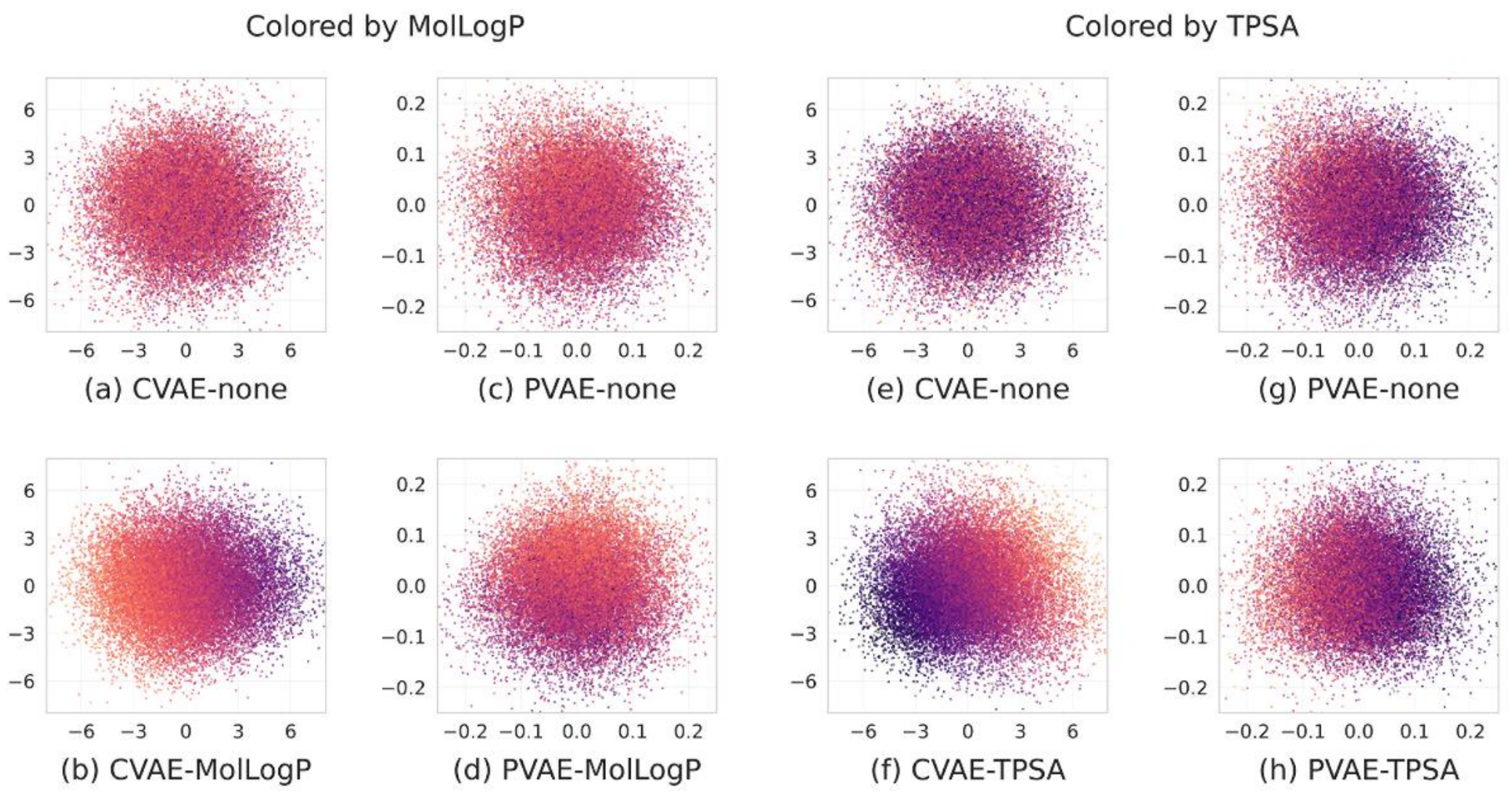

Fig. 1 2D-PCA visualizations of the latent spaces of various VAE models. Each point corresponds to one sample from the ZINC-250K test set. The points in (a)-(d) are colored by MolLogP, while the points in (e)-(h) are colored by TPSA

Next, we visualize how the target datasets are spread in the latent space with respect to the source dataset. It is interesting to note that in case of both types of PVAEs (with and without descriptor predictor modules) parts of the target datasets lay outside the source dataset according to the two principal components (Fig. 7).

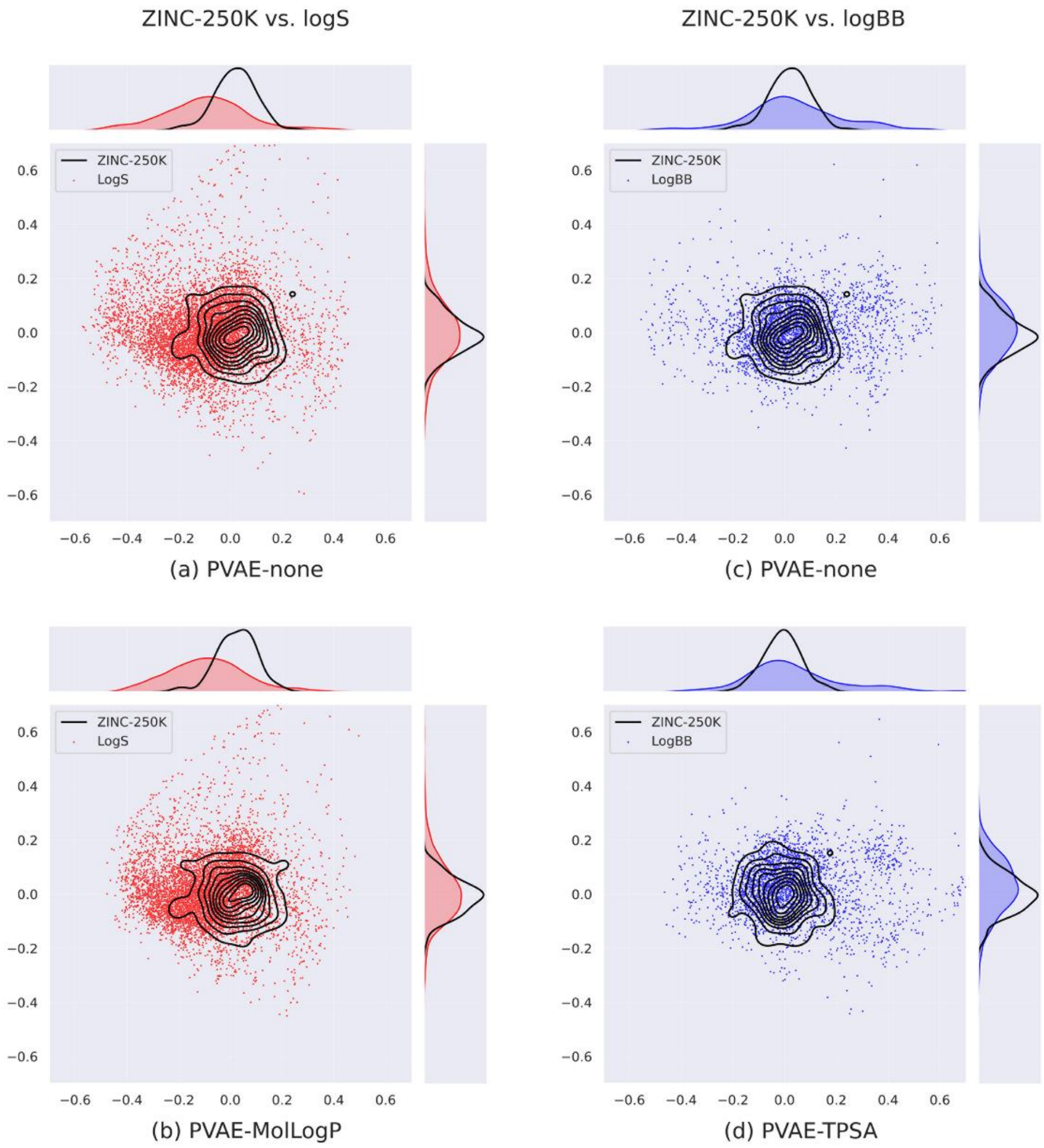

Fig. 7 2D-PCA visualizations of the latent spaces of PVAE models. Source dataset is in black, target datasets are in red and blue

To further explore how the distance from the source dataset affects the performance of the downstream QSAR models, we split the embeddings of the logS dataset into 10 clusters using K-Means clustering. For each cluster we compute the distance from the source dataset and the average

RMSE error of the prediction (note that we have predictions for all molecules, as each molecule belongs to the test split of one fold).

The results are shown in Fig. 8. The distance between the source dataset and the prior is 0.058. RMSE on the closest clusters (KL divergence < 0.1) is roughly between 0.5 and 0.7. RMSE for further clusters (KL divergence > 0.15) is larger than 0.8 with one exception: there is one cluster far from the prior (KL divergence = 0.22) and very low RMSE (0.43). Manual inspection shows that this "easy" cluster contains very small molecules mostly having less than 10 atoms. The choice of the number of clusters does not affect this conclusion, as shown in Additional file 1: Fig. S2.

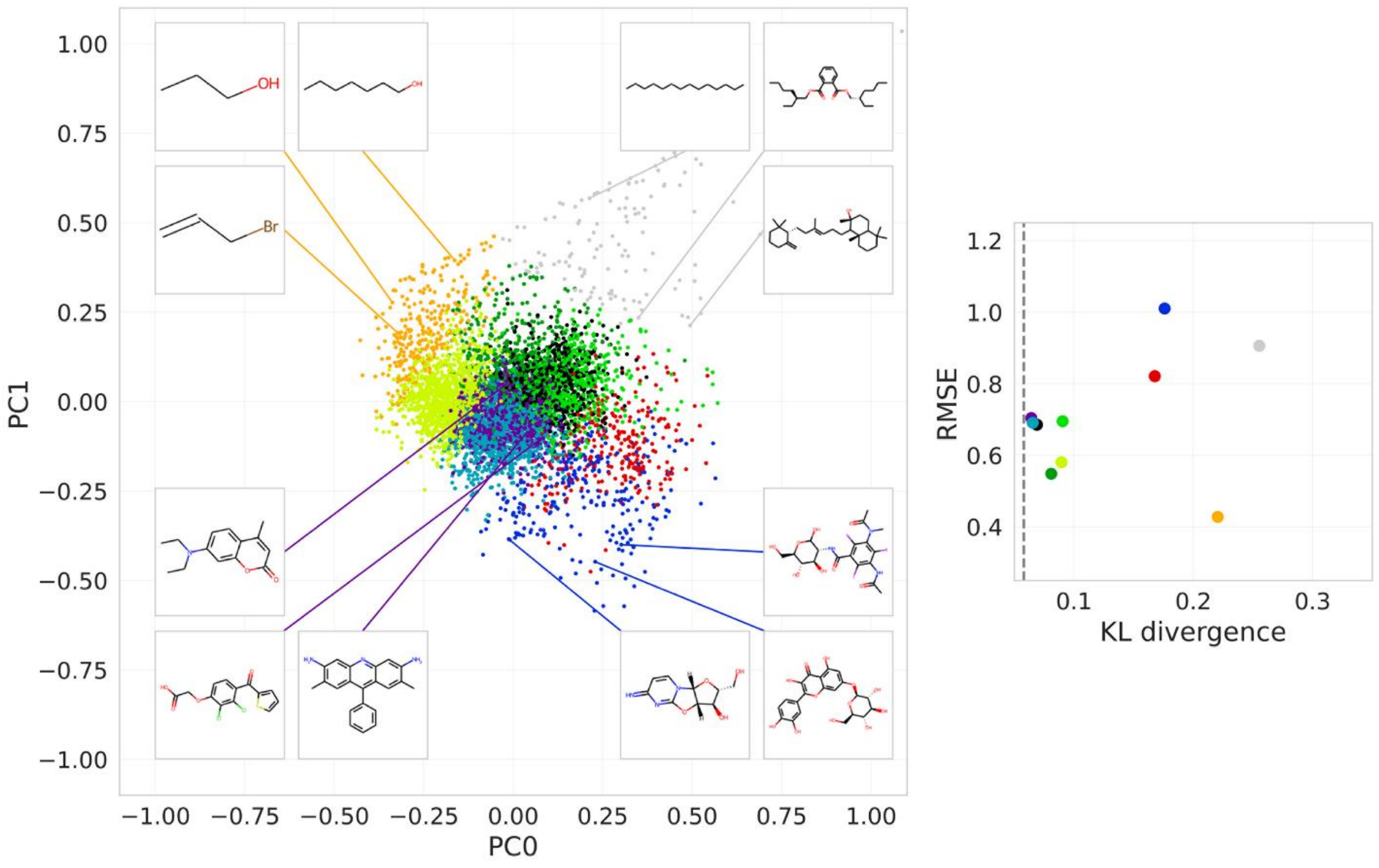

Fig. 2 On the left side, 2D-PCA visualization of the logS dataset is displayed on the latent space of a PVAE model trained jointly with a MolLogP predictor. The colors indicate the clusters found by the K-means algorithm. Few molecules from selected clusters are visualized. On the right side, the relation between average RMSE error and KL divergence of the clusters and the prior is shown. Each point corresponds to one cluster (using the same color map)

## Conclusions

Encoding molecules into fixed-sized continuous representations is crucially important for the development of physicochemical and biological property prediction algorithms. The main focus of this study was transferability of the embeddings extracted from VAEs rather than a comparison with the top machine learning approaches. We have demonstrated that the joint pre-training of VAE with the most relevant descriptor increased its predictive performance in the downstream tasks. We have also analysed the different sources of variance in the performance of the QSAR models.

We have further analysed the effect of using multiple descriptors at once. In some cases it improved downstream performance. We have shown that while more correlated descriptors tend to have stronger improvement on the performance of QSAR models, the correlation coefficient is not the sole factor affecting the performance.

We have seen that the size of the source dataset impacts reconstruction accuracy of the VAE, but does not affect downstream performance. We have confirmed that the joint pre-training of VAE with a descriptor predictor results in more smoothly organized latent space with respect to the selected descriptor.

Finally, we have identified that the source dataset ZINC-250K does not fully cover the target datasets in the space of molecular representations learned by VAE. The QSAR models we have analysed tend to have worse performance on the parts of the target datasets that are further from the source dataset, with an exception of a subset of very small molecules, indicating that choosing the source dataset in a way that the learned representations are diverse enough to cover target datasets might improve the performance of downstream QSAR models.

## List of abbreviations

QSAR: Quantitative Structure-Activity Relationship

QSPR: Quantitative Structure-roperty Relationship

ML: Machine learning

SMILES: Simplified Molecular Input Line Entry Specification

VAE: Variational Autoencoder

CVAE: Chemical Variational Autoencoder

logS: Aqueous solubility

logD: lipophilicity

logBB: blood-brain barrier penetration

PVAE: Penalized Variational Autoencoder

GRU: Gated Recurrent Units

MLP: Multilayer Perceptron

LR: Linear Regression

ResNet; Residual Network

RMSE: Root mean squared error

KL: Kullback-Leibler

TPSA: Topological surface area

**Supplementary Information**

Additional file 1: Improving VAE based molecular representations for compound property prediction

File format: DOC (Microsoft Word)

Additional file 1: Table S1 Hyperparameters of CVAE (a) and PVAE (b) and size of the models (c) used for pre-training. Fig S1. Correlation Heatmaps between descriptors calculated by RDkit package and aqueous solubility log S (a), lipophilicity logD (b) and blood-brain penetration logBB (c) datasets. Table S2 Hyperparameters of the 1D ResNet architecture. **Fig. S2** Clusters of the molecules from logS dataset in the latent space of a PVAE jointly pre-trained with a MolLogP predictor. Clusters are obtained using K-Means algorithm with varying number of clusters. In all cases there is one outlier cluster, without which there is a correlation between RMSE error of logS prediction and KL divergence between the cluster and the prior distribution.

### Availability of data and materials

The datasets, codes/scripts required to reproduce the results of this article are available in our GitHub repository: https://github.com/znavoyan/vae-embeddings.

### Competing interests

The authors declare that they have no competing interests.

### Funding

This work was supported by the RA MES State Committee of Science, in the framework of the research project № 20TTCG-1F004.

### Authors' contributions

Z. N., N. B. and H.S. conceptualized and supervised the project. A.T. carried out all the machine learning studies and descriptors calculations. L.K., H.K. and Z.N. formulated machine learning strategies. H.K. supervised their implementation. G.T. and L.A. collected all data. L.K., N.B. and H.K. prepared original manuscript. L.K., H.K., H. S., N.B. and Z. N. reviewed and edited it. All authors read and approved the final manuscript.